\documentclass[letterpaper, 10 pt, conference]{ieeeconf}  

\IEEEoverridecommandlockouts                              

\usepackage{cite}
\usepackage{amsmath,amssymb,amsfonts}
\usepackage{graphicx}
\graphicspath{{figures/}}
\usepackage{textcomp}
\usepackage{xcolor}
\usepackage{booktabs}
\usepackage{colortbl}

\usepackage{algorithm}
\usepackage{algpseudocode}

\def\BibTeX{{\rm B\kern-.05em{\sc i\kern-.025em b}\kern-.08em
    T\kern-.1667em\lower.7ex\hbox{E}\kern-.125emX}}
\begin{document}

\title{\Large \bf Kinematics-Aware Latent World Models for Data-Efficient Autonomous Driving\\
}


\author{Jiazhuo Li, Linjiang Cao, Qi Liu, and Xi Xiong
\thanks{J. Li, L. Cao, Q. Liu and X. Xiong are with the Key Laboratory of Road and Traffic Engineering, Ministry of Education, Tongji University, Shanghai, China, emails: 2534457@tongji.edu.cn, 2431743@tongji.edu.cn, liu\_qi@tongji.edu.cn, xi\_xiong@tongji.edu.cn.}
}
\maketitle

\begin{abstract}
Data-efficient learning remains a central challenge in autonomous driving due to the high cost and safety risks of large-scale real-world interaction. Although world-model-based reinforcement learning enables policy optimization through latent imagination, existing approaches often lack explicit mechanisms to encode spatial and kinematic structure essential for driving tasks.
In this work, we build upon the Recurrent State-Space Model (RSSM) and propose a kinematics-aware latent world model framework for autonomous driving. Vehicle kinematic information is incorporated into the observation encoder to ground latent transitions in physically meaningful motion dynamics, while geometry-aware supervision regularizes the RSSM latent state to capture task-relevant spatial structure beyond pixel reconstruction. The resulting structured latent dynamics improve long-horizon imagination fidelity and stabilize policy optimization.
Experiments in a driving simulation benchmark demonstrate consistent gains over both model-free and pixel-based world-model baselines in terms of sample efficiency and driving performance. Ablation studies further verify that the proposed design enhances spatial representation quality within the latent space. These results suggest that integrating kinematic grounding into RSSM-based world models provides a scalable and physically grounded paradigm for autonomous driving policy learning.

\end{abstract}

\section{INTRODUCTION}

Despite rapid progress in autonomous driving, achieving reliable decision-making in long-tail and safety-critical scenarios remains a fundamental challenge \cite{kuznietsov2024explainable}. Although recent systems demonstrate strong performance within restricted operational domains, failures still occur under distribution shift, including rare road user behaviors and extreme environmental conditions \cite{xu2025wod}.
Reinforcement learning (RL) provides a principled framework for sequential decision-making through interaction-driven optimization \cite{zhao2024survey}. However, its application to autonomous driving is severely constrained by data efficiency. Learning robust driving policies typically requires massive environment interactions, yet large-scale real-world data collection is costly, time-consuming, and potentially unsafe. While high-fidelity simulators reduce safety risks, they remain computationally expensive, often requiring millions of interaction steps for policy convergence. This interaction bottleneck fundamentally limits the scalability of pure model-free RL in complex driving environments.


To address the interaction bottleneck, world models (WM) learn compact latent representations of environment dynamics, enabling policy optimization through imagination rollouts instead of repeated real-environment interaction. By internalizing transition dynamics in latent space, WMs significantly improve sample efficiency and support long-horizon reasoning.
Building on this paradigm, we propose a task-relevant world-model framework tailored for autonomous driving. Unlike purely generative latent modeling, our approach explicitly aligns latent dynamics with spatial and kinematic structures critical for driving tasks. Visual observations are fused with vehicle kinematic states to construct a kinematics-grounded latent representation of the driving scene, while structured spatial supervision regularizes latent transitions toward geometrically consistent and driving-relevant semantics. Policy learning is then performed via imagination rollouts in the structured latent space, enabling data-efficient and dynamically coherent decision-making with reduced reliance on real-environment interaction.



Existing approaches to autonomous driving decision-making broadly fall into several paradigms. Model-based control has long served as a foundation, where optimization-based methods such as model predictive control (MPC) explicitly exploit system dynamics and constraints for trajectory planning and tracking \cite{falcone2007predictive}. In contrast, model-free RL learns policies directly from interaction data, with algorithms such as Proximal Policy Optimization (PPO) \cite{schulman2017proximal} and Soft Actor-Critic (SAC) \cite{haarnoja2018soft} enabling stable policy optimization.
Bridging control and learning, planning with learned dynamics integrates neural networks with search or model-based rollouts \cite{silver2016mastering,ha2018recurrent}. To further improve data efficiency, WM learn compact latent representations of environment dynamics, supporting planning or imagination-based policy optimization. Representative approaches include PlaNet and Dreamer \cite{hafner2019learning,hafner2019dream,hafner2025mastering}, which perform behavior learning via latent imagination, as well as large-scale generative world models for interactive environment synthesis \cite{bruce2024genie}.
Building on these advances, world-model-based methods have been applied to autonomous driving. Prior works enhance robustness through semantic masking or representation alignment between raw observations and privileged features \cite{gao2024enhance,yang2025raw2drive}. However, existing approaches primarily focus on representation filtering or generative modeling, without explicitly constraining latent dynamics to maintain geometrically consistent and physically grounded structures essential for closed-loop vehicle control.

Motivated by these observations, we build upon the Recurrent State-Space Model (RSSM) \cite{hafner2025mastering} and propose a task-relevant world-model framework that explicitly enforces spatial and kinematic consistency in latent dynamics for autonomous driving. Rather than treating latent representations as purely generative abstractions, our approach aligns the latent transition with driving-relevant spatial structure.
Specifically, we introduce two complementary mechanisms. First, low-dimensional vehicle kinematic states obtained from onboard sensing are incorporated into the observation encoder to ground latent transitions in physically meaningful motion dynamics. Second, structured spatial supervision is imposed through auxiliary prediction heads that estimate lane-relative geometry and neighboring vehicle states. The resulting gradients regularize the RSSM latent dynamics to preserve geometrically coherent and interaction-aware representations, thereby shaping the world model toward task-critical spatial structure.







The contributions of this paper are summarized as follows:
\begin{enumerate}
    \item We propose a kinematics-grounded world-model framework for autonomous driving that explicitly aligns latent dynamics with decision-critical spatial and motion structure.

    \item We introduce kinematic grounding and geometry-aware spatial regularization into RSSM-based latent transition learning, guiding the latent dynamics toward physically meaningful and interaction-aware representations.

    \item We empirically demonstrate significant improvements in data efficiency and driving performance, together with enhanced prediction accuracy and imagination fidelity of the learned latent dynamics.
\end{enumerate}

The remainder of this paper is organized as follows. Section II explains our modeling problems and the preliminary knowledge. Section III details our proposed framework. Section IV introduces the experimental setup and analyzes the results. Section V concludes the paper and discusses future directions.
\section{Preliminaries} \label{02_modeling}
This section describes the problem we studied and preliminary
knowledge of world models.
In real-world driving, an ego vehicle cannot access the complete and exact state of the environment, such as the precise intentions of other drivers or occluded road regions. Instead, it must rely solely on its onboard sensors, which provide partial and noisy observations. We model the autonomous driving task as a Partially Observable Markov Decision Process (POMDP).  The POMDP is defined by the tuple $(\mathcal{S}, \mathcal{A}, \mathcal{O}, \mathcal{T}, \mathcal{G}, \mathcal{R}, \gamma)$, where  $\mathcal{S}$ represents the true but unobservable environment state. $\mathcal{A}$ is the action space consisting of continuous steering and throttle/brake commands $\mathbf{a}_t = (a_{\text{steer}}, a_{\text{throttle}}) \in [-1, 1]^2$. $\mathcal{O}$ is the observation space. At each time step $t$, the ego vehicle receives an observation $\mathbf{o}_t$ composed of two modalities:
\begin{equation}
\mathbf{o}_t = \{I_t, \mathbf{v}_t\},
\end{equation}
where $I_t \in \mathbb{R}^{H \times W \times 3}$ is an image from a front-facing camera, and $\mathbf{v}_t \in \mathbb{R}^5$ is a vector of vehicle physics, including speed, steering angle, previous actions and yaw rate. Both of them can be directly obtained through raw sensors or physical calculations. $\mathcal{T}(s' \mid s, a): \mathcal{S} \times \mathcal{A} \times \mathcal{S} \rightarrow [0, 1]$ defines the conditional transition probability distribution over next states. $\mathcal{G}(o \mid s, a): \mathcal{S} \times \mathcal{A} \times \mathcal{O} \rightarrow [0, 1]$ defines the probability of observing given state and action. $\mathcal{R}: \mathcal{S} \times \mathcal{A} \rightarrow \mathbb{R}$ is the reward function balancing progress, safety, and rule compliance, while $\gamma \in [0, 1]$ is the discount factor. The agent's objective is to learn a policy $\pi(a_t | o_{\leq t})$ that maximizes the expected cumulative discounted reward:
\begin{equation}
\mathbb{E}_{\pi} \left[ \sum_{t=0}^{\infty} \gamma^t r_t \right].
\end{equation}
Due to partial observability in driving, the true transition dynamics $\mathcal{T}$ are unknown. To address this challenge, world-model-based reinforcement learning approximates belief dynamics using a learned latent representation. Instead of explicitly maintaining a probability distribution over environment states, the agent learns a compact latent variable $\hat{s}_{t} \in \mathcal{S}_{\text{latent}}$ that summarizes past information.

The world model usually includes the following components: a representation model $p(\hat{s}_{t} \mid \hat{s}_{t-1}, a_{t-1}, o_{t})$ that maps high-dimensional observations into a compact latent state; a latent transition model $p(\hat{s}_{t+1} \mid \hat{s}_t, a_t)$ that predicts future latent states; an observation model that reconstructs or predicts observations $p(\hat{o}_{t} \mid \hat{s}_{t})$; and a reward model $p(\hat{r}_{t} \mid \hat{s}_{t}, a_t)$ for value estimation. In some tasks, there also has a continue model $p(\hat{c}_{t} \mid \hat{s}_{t}, a_t) $ that predicts termination signal. Through this learned latent dynamics model, the agent can perform imagination rollouts entirely in latent space, enabling policy optimization without requiring real environment interaction at every step.
\section{The Kinematics-Aware Latent World Models}
\label{03_greedy_algorithm}

This section provides a detailed introduction to our proposed kinematics aware world model, which integrates multi-modal encoding, RSSM-based dynamics with driving-specific supervision, and latent-space policy learning to enable data-efficient training with explicit driving semantics.

\subsection{Multi-modal Encoding}
To address the limitations of pure image input, we propose an enhanced input representation that fuses image features with 5-dimensional vehicle physical information. These physical states can be accurately and efficiently obtained from onboard sensors (e.g., IMU, odometry), offering a reliable and low-cost alternative to inferring them from pixels.

Specifically, the image encoder processes the front-facing camera image \( I_t \) through a convolutional neural network (CNN):\( f_{\text{img}} = \text{CNN}(I_t; \theta_{\text{img}}) \). The physics encoder processes the normalized vehicle physics vector \( v_t \) through a multi-layer perceptron (MLP): \( f_{\text{phys}}= \text{MLP}(v_t; \theta_{\text{phys}}) \). Finally, the visual and physics features are concatenated to form the unified observation embedding:
\begin{equation}
e_t = \text{Concat}(f_{\text{img}}, f_{\text{phys}}) \in \mathbb{R}^{d_{\text{img}} + d_{\text{phys}}}.
\end{equation}

By incorporating this information explicitly, the world model does not have to infer dynamics solely from visual observations, allowing it to focus on learning environmental dynamics and interactions.

\subsection{Latent Dynamics Modeling}
The encoded observations embedding is then fed into a Recurrent State-Space Model (RSSM)\cite{hafner2025mastering}. At each time step, the model maintains a deterministic hidden state $h_t$ that summarizes past information, and a stochastic state $z_t$ that captures uncertainty. The transition is defined as:
\begin{equation}
\begin{aligned}
&h_t = f_\theta(h_{t-1}, z_{t-1}, a_{t-1}), \\
&\text{Prior: } \hat{z}_t \sim p_\theta(\hat{z}_t \mid h_t), \\
&\text{Posterior: } z_t \sim q_\theta(z_t \mid h_t, e_t),
\end{aligned}
\end{equation}
where $f_\theta$ is a recurrent neural network. $z_t$ is sampled from the posterior distribution during training, and prior distribution during inference. $(h_{t}, z_{t})$ together form the latent state. 

The basic world model loss consists of prediction loss and KL regularization loss:
\begin{equation}
\mathcal{L}_{\text{basic}} = \mathbb{E}_{q_\theta}\left[\sum_{t=1}^{T} \left(\mathcal{L}_{\text{pred},t} + \mathcal{L}_{\text{KL},t}\right)\right],
\end{equation}
where $\mathcal{L}_{\text{pred},t}$ includes the negative log-likelihood of observation reconstruction $-\ln p_\theta(\hat{o}_t \mid h_t, z_t)$, reward prediction $-\ln p_\theta(\hat{r}_t \mid h_t, z_t)$, and termination signal prediction $-\ln p_\theta(\hat{c}_t \mid h_t, z_t)$. The $\mathcal{L}_{\text{KL}, t} = D_{\text{KL}}(q_\theta(z_t \mid h_t, e_t) \parallel p_\theta(\hat{z}_t \mid h_t))$ is the KL divergence between the posterior and prior distribution, constraining the encoder to avoid extracting irrelevant information from observations and enhancing the consistency and predictability of the latent state.

\subsection{Driving-Specific Supervision Heads}
Relying solely on pixel reconstruction often neglects structured semantic information and cannot guarantee geometric consistency required for long-horizon interaction prediction, as critical driving elements such as lane boundaries and surrounding vehicles occupy only a small fraction of the visual input. We extend the output heads of the basic world model by adding two task-specific detection heads. The gradients of these two new heads are backpropagated to the world model, guiding the latent state to explicitly focus on key driving scene information.

A Lane Detection Head is designed to predict key lane-related information from the latent state, which are three critical indicators for lane keeping in autonomous driving: 
\begin{equation}
\hat{l}_t = f_{\text{lane}}(h_t,z_t) = \left[\hat{d}_{\text{left}}, \hat{d}_{\text{right}}, \hat{\Delta}_{\text{heading}}\right] \in \mathbb{R}^{3},
\end{equation}
where $\hat{d}_{\text{left}}$, $\hat{d}_{\text{right}}$ are distances to left and right lane boundaries, and $\hat{\Delta}_{\text{heading}}$ is the heading angle difference relative to the lane.

A Vehicle Detection Head is designed to predict the key information of surrounding vehicles from the latent state, which is critical for collision avoidance:
\begin{equation}
\hat{n}_t = f_{\text{nbr}}(h_t,z_t) \in \mathbb{R}^{12},
\end{equation}
specifically, the head predicts a 12‑dimensional vector representing the states of up to three surrounding vehicles, with each vehicle characterized by four attributes: the relative positions and relative speed along the ego vehicle's longitudinal and lateral axes.

It is worth noting that these signals are used only during training as auxiliary supervision and does not need to be provided during testing. We normalize all target values to eliminate dimensional heterogeneity across outputs. Both heads employ symlog MSE loss: 
\begin{align}
\mathcal{L}_{\text{lane}} &= \tfrac{1}{T}\sum_{t}\|\text{symlog}(l_t) - \text{symlog}(\hat{l}_t)\|_2^2, \\
\mathcal{L}_{\text{nbr}} &= \tfrac{1}{T}\sum_{t}\|\text{symlog}(n_t) - \text{symlog}(\hat{n}_t)\|_2^2,
\end{align}
where $\text{symlog}(x) = \text{sign}(x)\ln(|x|+1)$. The total training loss \( \mathcal{L}_{\text{total}} \) of the world model is then defined as:
\begin{equation}
\mathcal{L}_{\text{total}} = \mathcal{L}_{\text{basic}} + \mathcal{L}_{\text{lane}} + \mathcal{L}_{\text{nbr}}.
\end{equation}

\begin{figure}[htbp]
    \centering  
    \includegraphics[width=1.0\columnwidth]{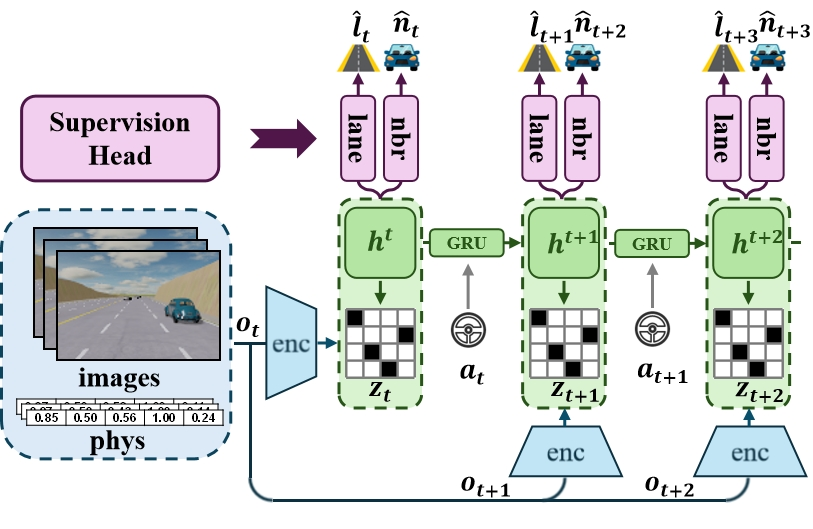}  
    \caption{World Model Learning. The network encodes multi-modal inputs into latent states and decodes multiple task-specific outputs including reconstruction, prediction, and driving-aware supervision signals.} 
    \label{fig:world_model_learning} 
\end{figure}

\subsection{Actor-Critic learning}
Same as DreamerV3~\cite{hafner2025mastering}, we train an actor network $\pi(a \mid \phi)$ and critic network $V(\phi)$ using imagined trajectories:

The critic estimates state values using $\lambda$-returns:
\begin{equation}
\mathcal{L}_{\text{critic}} = \mathbb{E}_{\tau^{\text{imag}}} \left[ \sum_{\tau=t}^{t+H-1} \frac{1}{2} \left( V(\phi_\tau) - R_\tau^\lambda \right)^2 \right],
\end{equation}
with the $\lambda$-return defined as:
\begin{equation}
R_\tau^\lambda = r_\tau + \gamma \left[ (1-\lambda) v_\psi(s_{\tau+1}) + \lambda R_{\tau+1}^\lambda \right],
\end{equation}
where $\gamma$ is the discount factor, $\lambda \in [0,1]$ is the trace decay parameter, $r_\tau$ is the immediate reward, $v_\psi(s_{\tau+1})$ is the bootstrap value estimate from the target network, and $H$ denotes the length of imagined trajectories. Intuitively, $\lambda$-return considers cumulative rewards within finite-length imagined trajectories, while using $v_\psi(s_n)$ at the endpoint to approximate future returns infinitely far ahead, thereby providing a global value estimate for each state.

We adopt the dynamics gradient mode, which directly maximizes the value function along the imagined trajectory. The actor loss is defined as the negative expected advantage, weighted by the cumulative product of discounts and continuation probabilities:
\begin{equation}
\mathcal{L}_{\text{actor}} = -\mathbb{E}_{\tau^{\text{imag}}} \left[ \sum_{\tau=t}^{t+H-1} w_{\tau} \cdot \left( R_\tau^\lambda - V(\phi_{\tau}) \right) \right],
\label{eq:actor_loss}
\end{equation}
where $w_{\tau} = \gamma^{\tau-t} \prod_{k=t}^{\tau-1} c_{k}$ is the cumulative weight, $c_{k}$ is the predicted continuation probability from the continuation head.
\begin{figure}[htbp]
    \centering  
    \includegraphics[width=1.0\columnwidth]{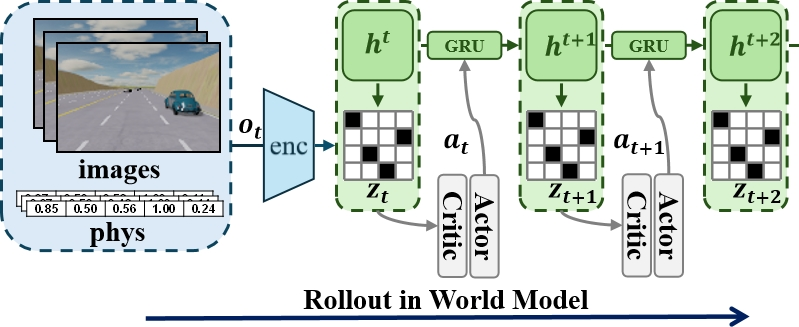}  
    \caption{Actor-Critic Learning. Imagined trajectories generated by the world model enable policy optimization in latent space. The actor predicts actions $\pi(a \mid \phi)$ while the critic estimates values $V(\phi)$ via $\lambda$-returns, allowing gradient-based updates without real environment interaction.} 
    \label{fig:actor_critic_learning} 
\end{figure}
\subsection{Reward Design}
The reward function $R$ in our driving task is composed of four components that balance progress, safety, and rule compliance:

The forward distance reward $R_1$ encourages making progress along the lane centerline:
\begin{equation}
R_1 = \beta_d \times (s_t - s_{t-1}) \times \mathbf{1}_{\text{pos}},
\end{equation}
where $s_t$ is the longitudinal projection of the vehicle onto the lane centerline, and $\mathbf{1}_{\text{pos}}$ indicates correct driving direction.

The speed reward $R_2$ encourages maintaining an appropriate speed:
\begin{equation}
R_2 = \beta_s \times (v / v_{\max}) \times \mathbf{1}_{\text{pos}},
\end{equation}
where $v$ is the current speed and $v_{\max}=80\,\mathrm{km/h}$.

The lane center offset penalty $R_3$ penalizes deviation from the lane center:
\begin{equation}
R_3 = -\beta_c \times (|y_t|) / w_{\text{lane}},
\end{equation}
where $y_t$ denotes lateral offset and $w_{\text{lane}}$ is lane width.

The termination reward/punishment $R_4$ provides a sparse terminal reward for completing the route or penalty for violations:
\begin{equation}
R_4 = 
\begin{cases}
r_{\text{success}} & \text{if episode completes successfully} \\
-p_{\text{crash}} & \text{if collision occurs} \\
-p_{\text{out}} & \text{if vehicle drives out of road} \\
\end{cases}.
\end{equation}

The total reward for each timestep is:
\begin{equation}
R = R_1 + R_2 + R_3 + R_4.
\end{equation}

\begin{algorithm}[t]
\caption{Kinematics-Aware World Model}
\label{alg:driveworld}
\begin{algorithmic}[1]
\Require environment, initial policy, hyperparameters
\Ensure trained policy
\State Initialize replay buffer $\mathcal{D}$
\State Initialize world model parameters $\theta$, actor parameters $\phi$, critic parameters $\psi$
\State Prefill $\mathcal{D}$ with random actions
\For{each iteration $k = 1, \dots, K$}
    \State Sample batch $\{(o_t, a_t, r_t)\}_{t=1}^{T}$ from $\mathcal{D}$
    \State \textcolor{gray}{\// World model learning}
    \State Encode $\mathbf{e}_t \leftarrow \textsc{Concat}\left(\textsc{CNN}(I_t; \theta_{\text{img}}), \textsc{MLP}(\mathbf{v}_t; \theta_{\text{phys}})\right)$
    \State Compute posterior and prior states via RSSM
    \State Compute losses: $\mathcal{L}_{\text{basic}}, \mathcal{L}_{\text{lane}}, \mathcal{L}_{\text{nbr}}$
    \State Update $\theta$ by minimizing total loss
    \State \textcolor{gray}{\// Behavior learning in imagination}
    \State Sample start states from posterior
    \State Generate imagined trajectory of length $H$ using actor and RSSM prior
    \State Compute $\lambda$-returns and advantages
    \State Update critic $\psi$ via regression to targets
    \State Update actor $\phi$ via dynamics gradient
    \State \textcolor{gray}{\// Environment interaction (parallel)}
    \State Rollout policy in environments for $N$ steps, adding to $\mathcal{D}$
\EndFor
\end{algorithmic}
\end{algorithm}
\section{Numerical Results} 
\label{04_experiments}

This section provides a detailed introduction to our experimental setup, model configuration, comparative experiment, ablation study and analysis of related results.

\subsection{Experimental Setup}
Our experiments are conducted in the MetaDrive\cite{li2022metadrive} autonomous driving simulation environment. The map we selected featuring multi-lane roads, moderate traffic density, and a mix of straight and curved segments, ensuring the task is sufficiently challenging while remaining tractable.

Key environment parameters include a straight and curve map string, traffic density of 0.1, action repeat of 20, image resolution of $120 \times 80$ pixels, time limit of 6000 environment steps, forward distance reward $\beta_d = 1.0$, speed reward $\beta_c = 0.1$, center offset penalty $\beta_c = 1.0$, crash and out of road penalty $p_{\text{crash}},p_{\text{out}} = 40.0$.

\subsection{Model Configurations}
The world model follows the DreamerV3 architecture with the enhancements described in Section~3. The CNN encoder and decoder employ a depth of 32, kernel size of 4, minimum resolution of 4, and SiLU activations. The MLP consists of 2 layers with 256 units, SiLU activations, and layer normalization. The RSSM uses a deterministic state size of 512, stochastic state size of $32 \times 32$, and recurrent depth of 1. Both actor and critic networks are configured with 2 layers of 512 units and layer normalization.

We use a batch size of 16 sequences, each of length 64. The learning rate is set to $1 \times 10^{-4}$. Optimization is performed using Adam. The imagination horizon is set to $H = 15$ steps. For GAE, we use discount factors $\gamma = 0.997$ and $\lambda = 0.95$. The KL regularization applies free bits of $1.0$, with dynamics scale of $0.5$ and representation scale of $0.1$. All models are trained for 1.6 million environment steps (equivalent to 80,000 interactive steps due to action repeat 20). 

\subsection{Comparative Experiments}
To demonstrate sample efficiency, we compare our framework against a standard model-free baseline: PPO implemented in Stable-Baselines3. PPO uses the same observation data, encoder and Actor Critic architecture of the same scale, and the same Metadrive configuration. PPO trains for 300000 agent steps with default hyperparameters: learning rate $2.5 \times 10^{-4}$, $n_{\text{steps}} = 256$, batch size 64, and 10 epochs per update.

\begin{figure}[htbp]
    \centering  
    \includegraphics[width=0.9\columnwidth]{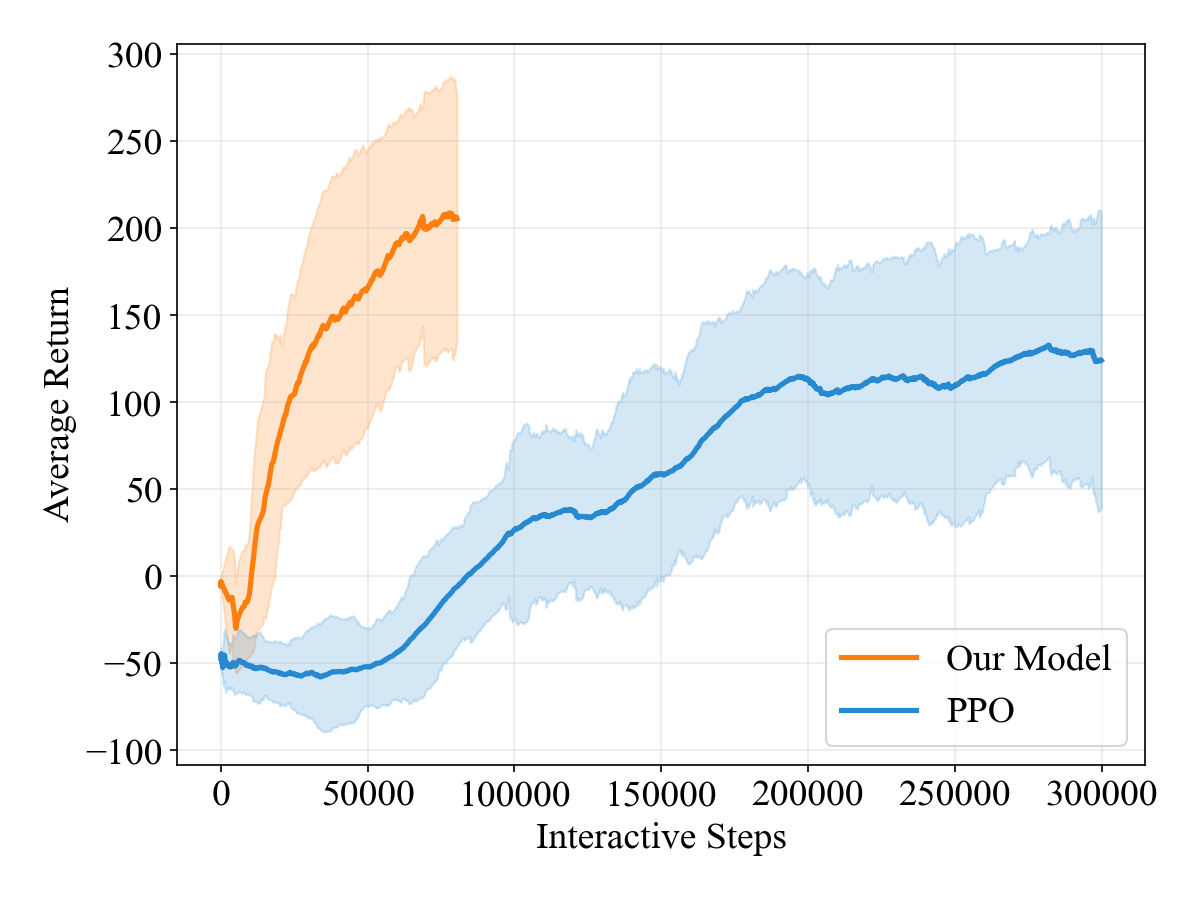}  
    \caption{A comparison between our model with PPO. The solid lines represent the averaged return, while the shaded area indicates the variability around the mean.} 
    \label{fig:train_return_vs_ppo} 
\end{figure}

The results in Figure~\ref{fig:train_return_vs_ppo} indicate that the world-model-based framework exhibits a relatively faster convergence rate. It reaches a stable high return (nearing 200) in just 80,000 real-environment steps. In contrast, PPO requires 300000 interaction steps to converge to a level below 150 scores.

\subsection{Ablation Studies}

\begin{figure}[t]
    \centering  
    \includegraphics[width=0.9\columnwidth]{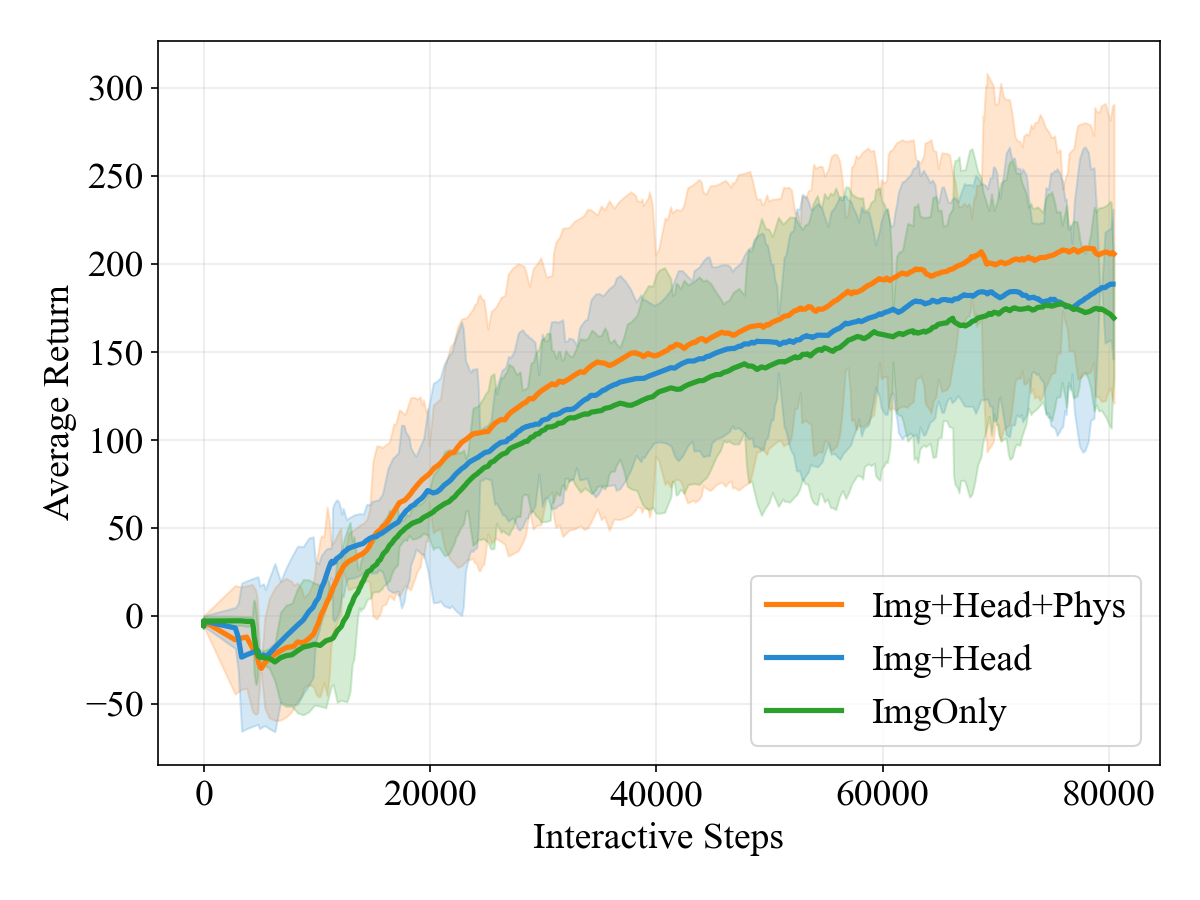} 
    \caption{Training curves of model variants in the ablation study. ImgOnly (green) uses images input alone; Img+Head (blue) adds lane and neighbor supervision heads; Img+Head+Phys (orange) further incorporates vehicle physics as input.} 
    \label{fig:Training curve} 
\end{figure}

We conducted a series of ablation experiments, and the results are presented in Figure~\ref{fig:Training curve} and Table~\ref{tab:ablation}. Our ablation study compares four model variants: ImgOnly, a baseline using solely images with reward and continuation heads; Img+Head, which adds lane and neighbor detection heads to image input; and Img+Head+Phys, our full framework integrating both multi-modal inputs and five supervision heads. Table~\ref{tab:ablation} also includes an experiment where the reward and continuation heads were removed.

The result indicate that adding the lane/neighbor heads to the image-only model improved the mean return (MR) by $9.7\%$ and the success rate (SR) by $16$ percentage points. After further incorporating physical information as input, MR continued to improve by $12.2\%$, with a total improvement of $23.1\%$. In addition, the reward and continuation heads played a crucial role in our experiment, and the performance of the model would significantly decline if they were removed.

These results validate that driving-specific supervision and multi-modal inputs are both critical, with their combination yielding synergistic improvements in driving performance.

\begin{table}[htbp]
\centering
\caption{Ablation Study Results}
\label{tab:ablation}
\begin{tabular}{@{}>{\bfseries}ccccccc@{}}
\toprule
\rowcolor{white}
\multicolumn{2}{c}{\textbf{Input}} & \multicolumn{3}{c}{\textbf{Heads}} & \multicolumn{2}{c}{\textbf{Metrics}} \\
\cmidrule(lr){1-2} \cmidrule(lr){3-5} \cmidrule(lr){6-7}
Image & Phys & Decoder & Rwd/Cont & Lane/Neigh & MR & SR \\
\midrule
\checkmark & $\times$ & \checkmark & \checkmark & $\times$ & 176.5 & 0.17 \\
\checkmark & $\times$ & \checkmark & \checkmark & \checkmark & 193.6 & 0.33 \\
\checkmark & \checkmark & \checkmark & $\times$ & \checkmark & 172.6 & 0.18 \\
\rowcolor{gray!20}
\checkmark & \checkmark & \checkmark & \checkmark & \checkmark & 217.2 & 0.49 \\
\bottomrule
\end{tabular}
\end{table}
\subsection{Synthetic Scenarios}
\begin{figure*}[t]
    \centering  
    \includegraphics[width=1.0\textwidth]{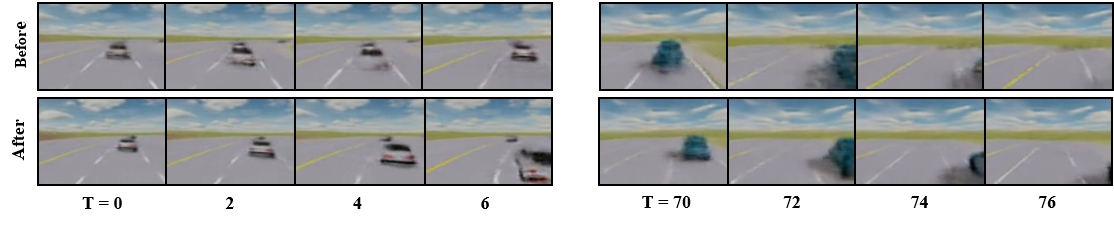}  
    \caption{Comparison of imagination quality across model variants. Top row: ImgOnly generates physically inconsistent rollouts with blurred vehicle positions (left) and confused lane markings (right). Bottom row: Img+Head+Phys produces stable, physically plausible predictions with correct semantic preservation of surrounding vehicles and lane markings.} 
    \label{fig:Imagined World} 
\end{figure*}
We compare the imagination quality of different model variants. As illustrated in Figure~\ref{fig:Imagined World}, the world model trained with only image input (ImgOnly) generates physically inconsistent rollouts. For example, when the ego car is preparing to overtake, the position of the preceding vehicle becomes blurred and undergoes abrupt, unrealistic shifts. In lane-changing scenarios, the model frequently confuses yellow solid lines with white dashed lines. In contrast, the combination of vehicle kinematics and auxiliary supervision effectively alleviates these problems. The imagined trajectories now maintain stable and physically plausible states for surrounding vehicles during interactions, and correctly preserve the color and type of lane markings during maneuvers, demonstrating improved physical grounding and semantic consistency in the latent space.

\section{CONCLUSIONS}
\label{05_conclusions}

This paper presents a task-relevant world-model framework for autonomous driving that improves latent representation learning through structured spatial supervision and kinematics-aware input. The proposed approach encourages the world model to capture driving-relevant spatial semantics instead of relying solely on visual reconstruction objectives. Simulation experiments demonstrate improved sample efficiency, faster convergence, and superior driving performance compared to baselines of the world-model only images and a model-free PPO agent, highlighting the benefits of explicitly structured supervision for safety-critical driving tasks.

Future work will focus on tighter integration of vehicle dynamics and world models to improve physical consistency in latent imagination. We also plan to extend the framework to offline learning settings, leveraging large-scale driving datasets to enhance data efficiency and safety. In addition, scaling the approach to multi-agent scenarios will be an important direction for modeling interactive traffic behaviors in complex environments.

\bibliographystyle{IEEEtran}
\bibliography{ssp}

\end{document}